\documentclass[lettersize,journal]{IEEEtran}
\usepackage{amsmath,amsfonts}
\usepackage{algorithmic}
\usepackage{algorithm}
\usepackage{array}
\usepackage[caption=false,font=normalsize,labelfont=sf,textfont=sf]{subfig}
\usepackage{textcomp}
\usepackage{stfloats}
\usepackage{url}
\usepackage{verbatim}
\usepackage{graphicx}
\usepackage{multirow}
\usepackage{booktabs}
\usepackage{cite}
\usepackage{hyperref}
\hypersetup{
    colorlinks=true,
}

\usepackage{amssymb}
\usepackage{pdfpages}
\usepackage{caption}
\usepackage[numbers]{natbib}

\begin{document}

\title{NeuroInspect: Interpretable Neuron-based Debugging Framework through Class-conditional Visualizations }

\author{Yeong-Joon Ju, Ji-Hoon Park, and Seong-Whan Lee,~\IEEEmembership{Fellow,~IEEE}
\thanks{This work was supported by the Institute of Information \& communications Technology Planning \& Evaluation (IITP) grant funded by the Korea government (MSIT) (No. 2019-0-00079, Artificial Intelligence Graduate School Program (Korea University) and No. 2022-0-00984, Development of Artificial Intelligence Technology for Personalized Plug-and-Play Explanation and Verification of Explanation). (\it{Corresponding author: Seong-Whan Lee.)}}

\thanks{Yeong-Joon Ju, Ji-Hoon Park, and Seong-Whan Lee are with the Department of Artificial Intelligence, Korea University, Seoul 02841, South Korea (e-mail: yj\_ju@korea.ac.kr; jhoon\_park@korea.ac.kr; sw.lee@korea.ac.kr). }}



\maketitle


\begin{abstract}
Despite deep learning (DL) has achieved remarkable progress in various domains, the DL models are still prone to making mistakes. This issue necessitates effective debugging tools for DL practitioners to interpret the decision-making process within the networks. However, existing debugging methods often demand extra data or adjustments to the decision process, limiting their applicability. To tackle this problem, we present NeuroInspect, an interpretable neuron-based debugging framework with three key stages: counterfactual explanations, feature visualizations, and false correlation mitigation. Our debugging framework first pinpoints neurons responsible for mistakes in the network and then visualizes features embedded in the neurons to be human-interpretable. To provide these explanations, we introduce CLIP-Illusion, a novel feature visualization method that generates images representing features conditioned on classes to examine the connection between neurons and the decision layer. We alleviate convoluted explanations of the conventional visualization approach by employing class information, thereby isolating mixed properties. This process offers more human-interpretable explanations for model errors without altering the trained network or requiring additional data. Furthermore, our framework mitigates false correlations learned from a dataset under a stochastic perspective, modifying decisions for the neurons considered as the main causes. We validate the effectiveness of our framework by addressing false correlations and improving inferences for classes with the worst performance in real-world settings. Moreover, we demonstrate that NeuroInspect helps debug the mistakes of DL models through evaluation for human understanding. The code is openly available at \href{https://github.com/yeongjoonJu/NeuroInspect}{https://github.com/yeongjoonJu/NeuroInspect}.
\end{abstract}


\section{Introduction}
Deep learning (DL) practitioners often seek to understand why a trained model makes specific mistakes, but the lack of accountability and transparency in DL models hinders the interpretation of their decision-making process. Thereby, DL practitioners increasingly require tools to aid in interpreting the underlying reasons for decisions of the model with the widespread adoption of DL models. However, determining the rationale behind the predictions of the model remains challenging. Furthermore, it is even more difficult to provide human-understandable explanations without additional data or changes to the decision process to be utilized as practical tools.

Existing explanation methods interpret the decision-making process in various ways to examine black-box models. Attribution-based explanation approaches~\cite{simonyan2014deep,bach2015pixel,selvaraju2017grad,sundararajan2017axiomatic,9961149,9457245} visualize the reasons for a decision at the input level by highlighting areas (i.e., tokens in a sentence or regions in an image) contributing to the prediction of the model. These approaches show the view of the model from the input, but such localized explanations can lead to a misleading understanding of the overall decision-making process~\cite{adebayo2018sanity,adebayo2020debugging,leavitt2020towards}. Concept-based methods~\cite{kim2018interpretability,ghorbani2019towards,yeh2020completeness,abid2022meaningfully} introduce a concept bank utilizing concept activation vectors (CAVs)~\cite{kim2018interpretability} to provide more human-understandable explanations; CAVs are vectors in the direction of activation for a set of examples of concepts. These concepts are selected by domain experts or acquired using an automatic framework that builds a concept library from the collected datasets. Such methods can transfer confirmation bias by humans or data bias in sets of examples. Our methods are based on the perspective that the decision-making process of the model is a procedure for understanding the input data (deep feature extractor) and making decisions (dense linear layer). In particular, the model embeds features and computes the probabilities of each class based on the values of the activated features. These features can be translated into human-understandable characteristics through feature visualization~\cite{nguyen2016multifaceted,olah2017feature,bau2017network} or neuron labeling~\cite{hernandez2021natural, oikarinen2022clip}. However, the existing feature visualization methods require data collection to probe multiple facets~\cite{nguyen2016multifaceted,goh2021multimodal} or generate images that are difficult for humans to understand \cite{olah2017feature}. In addition, neuron-labeling methods have the potential to convey incorrect explanations and confirmation bias because they allocate text to neurons using a network learned with human labels after probing a few examples activated for investigating neurons~\cite{hernandez2021natural,oikarinen2022clip}.

To reveal the underlying reasons for mistakes in the model, previous studies have employed diverse explanatory methods, such as the utilization of the concept bank and neuron interpretation. Based on the interpretation of decisions utilizing the concepts, CoCoX~\cite{akula2020cocox} and CCE~\cite{abid2022meaningfully} offer concept labels corresponding to false correlation in a counterfactual explanation manner, scoring concepts accountable for mistakes with a learned concept bank. However, they require reconstruction of the concept bank to acquire CAVs whenever the parameters of the network are changed. Furthermore, depending on the predefined concepts can be problematic for domain-specific tasks, such as fine-grained classification, as it may lead to incomplete explanations. Other approaches~\cite{wong2021leveraging, yuksekgonul2022post, liang2023multiviz} modify the decision-making process to enhance the interpretability of network decisions by sparsely tuning the decision layer. Post-hoc CBM~\cite{yuksekgonul2022post} projects embeddings from a network onto the concept subspace defined by CAVs and then predicts classes from their projections by leveraging the sparse linear layer as an interpretable predictor. Instead of scoring the defined concepts, \citet{wong2021leveraging} and \citet{liang2023multiviz} rank and visualize the top features (activated by neurons) for predictions by utilizing feature visualizations~\cite{yosinski2015understanding} and attribution-based explanations~\cite{lime}, respectively. These approaches to changing the decision-making process may degrade performance or provide explanations that are difficult to understand since neurons mainly encapsulate abstract features.


In this paper, we present NeuroInspect, a neuron-based debugging framework that interprets the underlying reasons for mistakes made by DL models without requiring additional data or modifying the decision-making process. NeuroInspect first indicates neurons that correspond to the main causes of mistakes in a counterfactual explanation manner. The framework then translates the features that activate neurons into human-understandable visualizations. However, the conventional feature visualization method~\cite{olah2017feature} generates images that are abstruse for humans to understand since a neuron may encompass multifaceted features. To address this issue, we introduce a class-conditional feature visualization method, CLIP-Illusion, which illustrates features manifested within a specific class, along with the constraints imposed by a web-scale multimodal model. The CLIP-Illusion creates images that instantiate abstract and mixed features contained in neurons into specific features expressed in a specific class, facilitating debugging of the model. Furthermore, since our framework directly pinpoints neurons mainly responsible for mistakes, NeuroInspect can mitigate false correlations learned from data without any additional training data. We show that our approach can identify and alleviate spurious correlations, leading to enhanced performance and more reliable decision-making by DL models. Because our approach does not require any changes to the data or networks, our neuron-based framework can be deployed as a practical model debugging method. The contributions of this study can be summarized as follows:
\begin{itemize}
    \item We present NeuroInspect, a neuron-based debugging framework that explains model mistakes without requiring additional data or network modifications, leading to a practical model debugging framework.
    \item We propose CLIP-Illusion, a method for generating human-understandable feature visualizations that enhance interpretability by considering the connections between the features and the decision layer. The CLIP-Illusion addresses the challenges associated with applying traditional visualization methods to explain model decisions.
    \item We demonstrate the effectiveness and practicality of our debugging framework through two experiments, which uncover the causes of mistakes and mitigate false correlations in a real-world setting.
\end{itemize}

\section{Related Works}
\subsection{Interpretation for Decision-making Process}

Concept Activation Vectors (CAVs)~\cite{kim2018interpretability} have been successfully adopted in several approaches~\cite{yuksekgonul2022post, oikarinen2022clip, abid2022meaningfully} to explain decision-making processes or units in a network using high-level concepts. These methods utilize user-defined or automatically discovered concepts from data to train linear classifiers in the bottleneck layers of the probed network~\cite{ghorbani2019towards, yeh2020completeness}. CoCoX~\cite{akula2020cocox} and CCE~\cite{abid2022meaningfully} employ a concept bank, a collection of CAVs, to interpret the decision-making process using a counterfactual explanation approach. The CCE method ensures the validity of counterfactual concepts by solving a constrained optimization problem. However, methods using the concept bank require the retraining of linear classifiers owing to each change or tuning in the feature extractor. In our work, we directly utilize embedded features rather than relying on the concept bank learned from external data while we employ a counterfactual explanation approach to interpret the errors in the model. Other studies by \citet{wong2021leveraging} and \citet{yuksekgonul2022post} convert the decision layer in the network into a sparse linear layer to improve the interpretability of predictions, visualizing the neurons or leveraging the concept bank. \citet{wong2021leveraging} first retrain the decision layer using elastic net regularization~\cite{zou2005regularization} and then visualize the features in the sparse linear layer. \citet{liang2023multiviz} presented a debugging framework to interpret the behaviors of vision-and-language models. However, their framework is difficult to adopt as a practical debugging framework because they offer uninterpretable explanations, such as the local analysis of a sample and coefficients for the connections of neurons to the decision layer.

\subsection{Network Interpretation}
\citet{erhan2009visualizing} optimize a randomly initialized input space to maximally activate specific units in the early layers of a network. Feature Visualization (FV)~\cite{olah2017feature} extends the work presented in \citet{erhan2009visualizing} to neurons, channels, and layers with various augmentations. We target interpreting channels or neurons in a penultimate layer directly connected with the decision layer, depending on the architecture of the layer. \citet{nguyen2016synthesizing, nguyen2017plug} introduce the generative model $G$ to produce realistic visualizations by identifying optimal latent codes in the GAN space. They update the latent code to generate the image that highly activates a given neuron instead of using a learned prior. \citet{simonyan2014deep} and \citet{yin2020dreaming} produce images from noise to maximize a score for a particular class by using an image prior and statistics trained in batch normalization, respectively. \citet{pmlr-v162-ghiasi22a} extend the works~\cite{simonyan2014deep, yin2020dreaming} based on various experiments for augmentations. Unlike the works~\cite{simonyan2014deep,yin2020dreaming,pmlr-v162-ghiasi22a} focusing on producing images for classes learned in a classifier, we aim to generate visualizations to represent attributes of a given neuron in related classes. \citet{ghiasi2022vision} visualize neurons in vision transformers by extending \citet{pmlr-v162-ghiasi22a}. \citet{bau2017network, bau2020understanding} assign pixel-wise semantic concepts with image segmentation labels after observing highly activated regions in exemplar images. \citet{hernandez2022natural} label a given neuron with texts generated by a network trained with top-activated exemplar images and human annotations. \citet{oikarinen2022clip} leverage CLIP networks~\cite{radford2021learning} to calculate the concept-activation matrix from a concept dataset, and then they match concept labels to given neurons based on similarity. Such approaches based on observing top-activated images may convey wrong biases by humans or probing datasets.

\begin{figure*}[]
\centering
\includegraphics[width = 1.\linewidth]{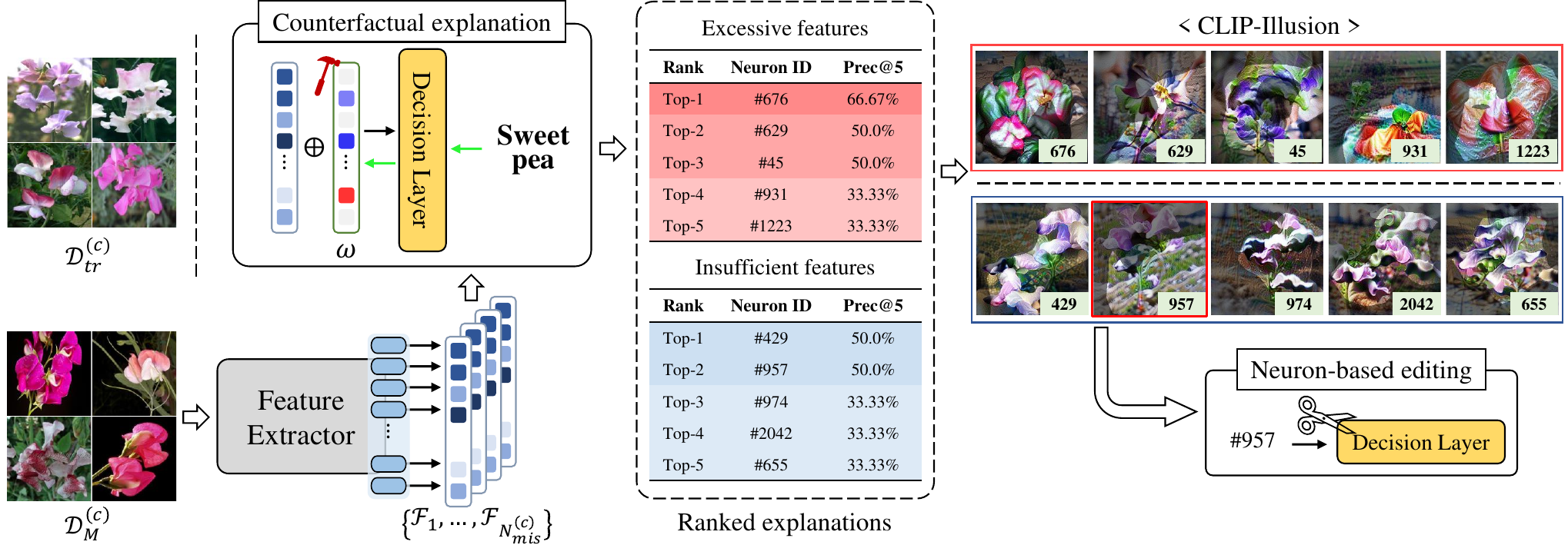}
\captionof{figure}{
    \textbf{Overall process of our debugging framework.} Given the mistake samples, we first acquire a set of features with the feature extractor of the model being analyzed. The counterfactual explanation adjusts the features with $\omega$ tuned to be predicted to a correct label. We then rank the neurons that make mistakes based on all $\omega$ values for the features. Finally, we provide visualizations of the top-ranked neurons, which are the main causes of mistakes across all mistake samples. We calculate rank rates for each neuron by extracting the top-5 neurons contributing to mistakes for each sample. After observing the visualizations, DL practitioners can improve the model performance using our editing method or other methods, such as data augmentation and architecture modifications.}
\label{fig:framework}
\end{figure*}

\section{NeuroInspect: Neuron-based Debugging Framework}

\subsection{Problem Statement}
In this section, we present the problem that our framework aims to address. We focus on object recognition tasks for supervised image classification. Given a training dataset $\mathcal{D}_{tr}=\{(x_1,y_1),...,(x_{N_{tr}}, y_{N_{tr}})\}$, where each sample consists of an input image $x$ and its corresponding class label $y$, the classification network $h$ learns to categorize an input image $x$ into the correct class $y$. The trained network $h$ can be decomposed into a feature extractor $h_b: \mathbb{R}^i \rightarrow \mathbb{R}^D$ and a decision layer $h_t: \mathbb{R}^D \rightarrow \mathbb{R}^C$, where $i$, $D$, and $C$ represent the dimensions of the input and output features $\mathcal{F}$, the number of classes, respectively. $\mathcal{F}$ denotes a vector composed of the features of image $x$.

Our primary objective is to provide human-understandable explanations for the main reasons behind the mistakes of network $h$ on the test dataset $\mathcal{D}_{ts}$, without requiring additional data or network modification. Given the trained network $h$ and the mistake samples $\mathcal{D}_M^{(c)} \subset \mathcal{D}_{ts}$ in class $c$, our framework explains the underlying reasons for the mistakes in class $c$. Moreover, we not only offer interpretations of the main causes of misclassifications but also aim to improve the predictions of the model by mitigating false correlations within class $c$. For instance, a `water jet' may be frequently associated with the class `fireboat' in the training data, but the presence of a `fireboat' does not necessarily imply the presence of a `water jet.' The spurious feature becomes harmful when the classifier considers the feature as a significant feature of the class, even in the absence of the class object in the image.

\subsection{Overall Process of NeuroInspect} 
Suppose a model makes mistakes $\mathcal{D}_M^{(c)}$ by predicting incorrect labels for some input images in a validation set. Given these images and their correct labels, our framework offers easy-to-understand visualizations of the features responsible for the mistakes, as shown in Fig.~\ref{fig:framework}. NeuroInspect first extracts a set of features $\mathcal{F}$ from $\mathcal{D}_M^{(c)}$ using the feature extractor $h_b$. To explain the reasons for the mistakes, we employ a counterfactual explanation approach modifying the features to make the model predict the correct label. We then acquire the neurons that mainly raise the mistakes by ranking them based on their tuned scores. To make these neurons more interpretable, we use CLIP-Illusion to generate visualizations that consider the connection to the decision layer by conditioning a correct class or classes correlated to features embedded in the neurons. After observing the explanations, DL practitioners can then modify the decision layer of the model to mitigate false correlations between neurons and the class $c$ using our editing method or leverage their comprehension of the causes of the mistakes to improve the model in other ways. Note that we only need to investigate a small number of neurons since we filter out irrelevant neurons using counterfactual explanations.

\begin{figure*}[]
\centering
\includegraphics[width=1.\linewidth]{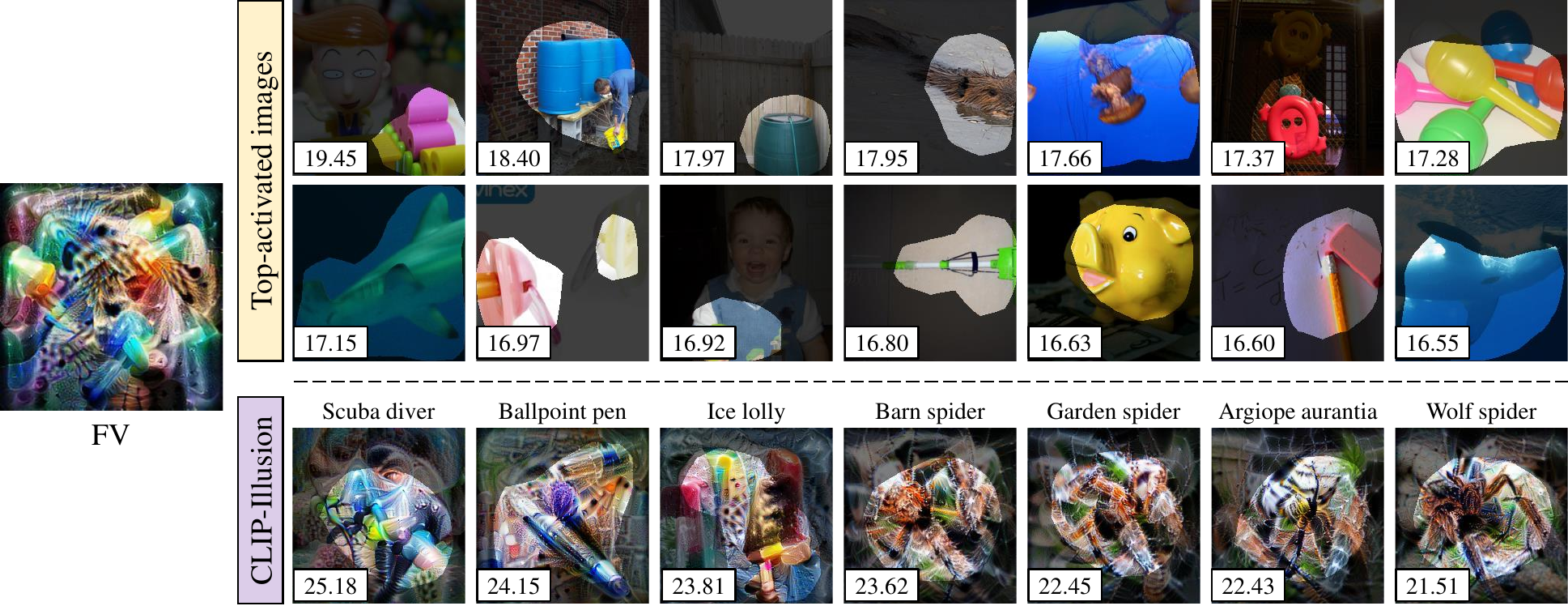}
\captionof{figure}{
    \textbf{Comparison of FV, CLIP-Illusion, and top-activated images.} This figure shows visualizations for neuron 3 of the penultimate layer in ResNet-18. The values at the bottom left of the image represent activation values for neuron 3. The top-activated images show examples with the highest activations for neuron 3 in the validation set of the ImageNet dataset~\cite{5206848}. CLIP-Illusion illustrates representations that activate the neuron in each class, showing how the neuron responds to different class-specific features.}
\label{fig:fv_comp}
\end{figure*}

\subsection{Neuron-based Counterfactual Explanations}
Our approach to providing counterfactual explanations for model errors is inspired by the work of \citet{abid2022meaningfully}. We find out neurons responsible for model errors by adjusting their activations to increase the probability of the correct class. To flip the prediction into the target class via minimal variations for features, we employ elastic net regularization~\cite{zou2005regularization}, a generalization of LASSO and ridge regression, which constrains $\omega$ altering the features used to the decision. The $\omega\in\mathbb{R}^D$ denotes a vector initialized to zero values. We optimize $\omega$ by minimizing the following loss function:
\begin{equation}
    \mathcal{L}_{mis} = \mathcal{L}_\text{CE}(h_t(h_b(x)+\omega), y) + \lambda_1\left|\omega\right|_1 + \lambda_2\left|\omega\right|_2,
\end{equation}

where $\mathcal{L}_\text{CE}$ denotes the cross-entropy loss function used to train the classification model. $\lambda_1$ and $\lambda_2$ denote the hyperparameters for the regularization term, set to 0.1 and 0.01, respectively. Values of the vector $\omega$ indicate the degree of insufficiency or excessiveness of the activation for each feature to decide the target class, e.g., values of $\omega$ more than 0 for a given feature mean the activation of the feature is insufficient for predicting the correct class. Consequently, we rank all neurons contributing to mistakes in the model based on the values of $\omega$ for the mistake samples $\mathcal{D}^{(c)}_M$. We follow a structured approach to attain core neurons. First, we obtain a set of $\omega$ variants optimized for all mistake samples to ensure that the features are classified as the correct class. Each $\omega$ corresponds to an alteration in the features of a single sample that influences a change in its decision. Subsequently, the top 5 ranked neurons for each sample are selected based on the values of the elements in $\omega$. Following this, we calculate the top-5 precision for all neurons across all mistake samples $\mathcal{D}^{(c)}_M$. Neurons with higher precision activate features that contribute significantly to mistakes in the model. From the perspective of the network, these explanations are divided into two types of reasons: excessive features and insufficient features, as shown in Fig.~\ref{fig:framework}. Excessive features are the undesired activations that cause the decision layer to wrongly categorize the test samples, whereas insufficient features are the lack of activations concerning the ground-truth class. These neuron-based counterfactual explanations enable the interpretation of the underlying reasons for errors regardless of the domain, combining feature visualizations for the neurons via CLIP-Illusion.

\subsection{CLIP-Illusion: Class-conditional Feature Visualization Using CLIP Network}
FV~\cite{olah2017feature} synthesizes the images $I_{opt}$ for a given neuron through activation maximization from Gaussian noise. This method represents features that effectively activate a given neuron without probing the data. However, as FV illustrated in Fig. \ref{fig:fv_comp}, visualization often derives images that are difficult for humans to understand because the images are abstract and mixed with several characteristics of one neuron owing to multiple facets of the neuron~\cite{nguyen2016multifaceted}. In addition, DL practitioners cannot be aware of how such features are expressed in the images within each class. Interpreting the concepts of neurons by observing top-activated images is also dependent on the probing datasets, leading to interpretations that are biased toward the data distribution. As shown in the top-activated images in Fig.~\ref{fig:fv_comp}, we could only recognize some of the implied concepts of neuron 3, excluding the concept of spiders. To address these limitations, we visualize representations of a neuron for class $c$ by leveraging CLIP networks~\cite{radford2021learning}, which comprise two encoders trained on 400M images/sentence pairs from the web. To visualize representations of a neuron $n$ in a class $c$, we first aim to increase activation of $n$ and a logit for $c$ via optimizing the image $I_{opt}$ initialized from Gaussian noise by minimizing the following loss:
\begin{equation}
    \mathcal{L} = -(\alpha_n + \gamma l_c),
\end{equation}

where $\alpha_n$ and $l_c$ denote activation of the neuron $n$ and a logit for the class $c$, respectively. $\gamma$ is a hyperparameter to control representations for classes with a range of 0 to 1. The $\gamma$ is introduced to prevent excessive class activation from inhibiting the activation of the neuron from being investigated. However, the optimized images may not be clear enough to understand due to excessively activating $\alpha_n$ and $l_c$ without any constraints while containing characteristics of the neuron $n$ in the class $c$. Thus, we additionally employ the CLIP similarity $\mathcal{D}_{\text{CLIP}}$ to diminish the abstruse visualizations for human understanding as follows:
\begin{equation}
    \mathcal{D}_\text{CLIP} = D_{cos}(E_I(I_{opt}), E_T(T_c)),
\end{equation}

where $E_I$ and $E_T$ denote the image and text encoders, respectively. $D_{cos}$ corresponds to the cosine distance between the two embeddings. These networks possess the ability to map images and texts based on web-scale multi-modal knowledge. Thus, we imitate human understanding by optimizing the image $I_{opt}$ in alignment with text prompts $T_c$ for the class $c$. In the end, we generate human-understandable visualizations by optimizing $I_{opt}$ to minimize the following loss:
\begin{equation}
    \mathcal{L}_\text{CI} = -(\mathcal{D}_\text{CLIP}+\epsilon)\cdot(\alpha_n + \gamma l_c),
\end{equation}

where $\epsilon$ denotes a hyperparameter to prevent domain misalignment of CLIP. $I_{opt}$ is tuned to increase the activation of neuron $n$ for class $c$ while also maximizing the CLIP similarity between the generated image and text prompt $T_c$. By doing so, we generate class-conditional feature visualizations that emphasize the relevant representations of a neuron for a specific class and are also understandable to humans. Finally, we mask out regions with low activations for the neuron in the images to occlude the interference of other information except for the activating features in the generated images. This ensures that only the areas of the image that are relevant to the concept represented by the neuron are included in the final visualization. As illustrated in Fig.~\ref{fig:fv_comp}, the masked regions correspond to areas that embody properties that activate the neuron. DL practitioners can find out the neurons that respond to false correlations via the CLIP-Illusion from insufficient features in counterfactual explanations of mistakes. This approach is more flexible than relying solely on concept labels or text labels. For instance, as shown in Fig. 1, neurons 974, 2042, and 655 correspond to primary features that the model misunderstood, such as the shape of the flowers. This confusion stems from training on a dataset in which the shapes of flowers are predominantly represented. Although the samples in $D_{mis}^{(c)}$ contain clear flower traits, the model makes errors owing to overfitting certain floral shapes. On the other hand, defining such fine-grained features using concept or text labels can be challenging due to ambiguity. Therefore, NeuroInspect provides a domain-independent and flexible approach for model debugging.

\begin{figure*}[]
\centering
\includegraphics[width=1.\linewidth]{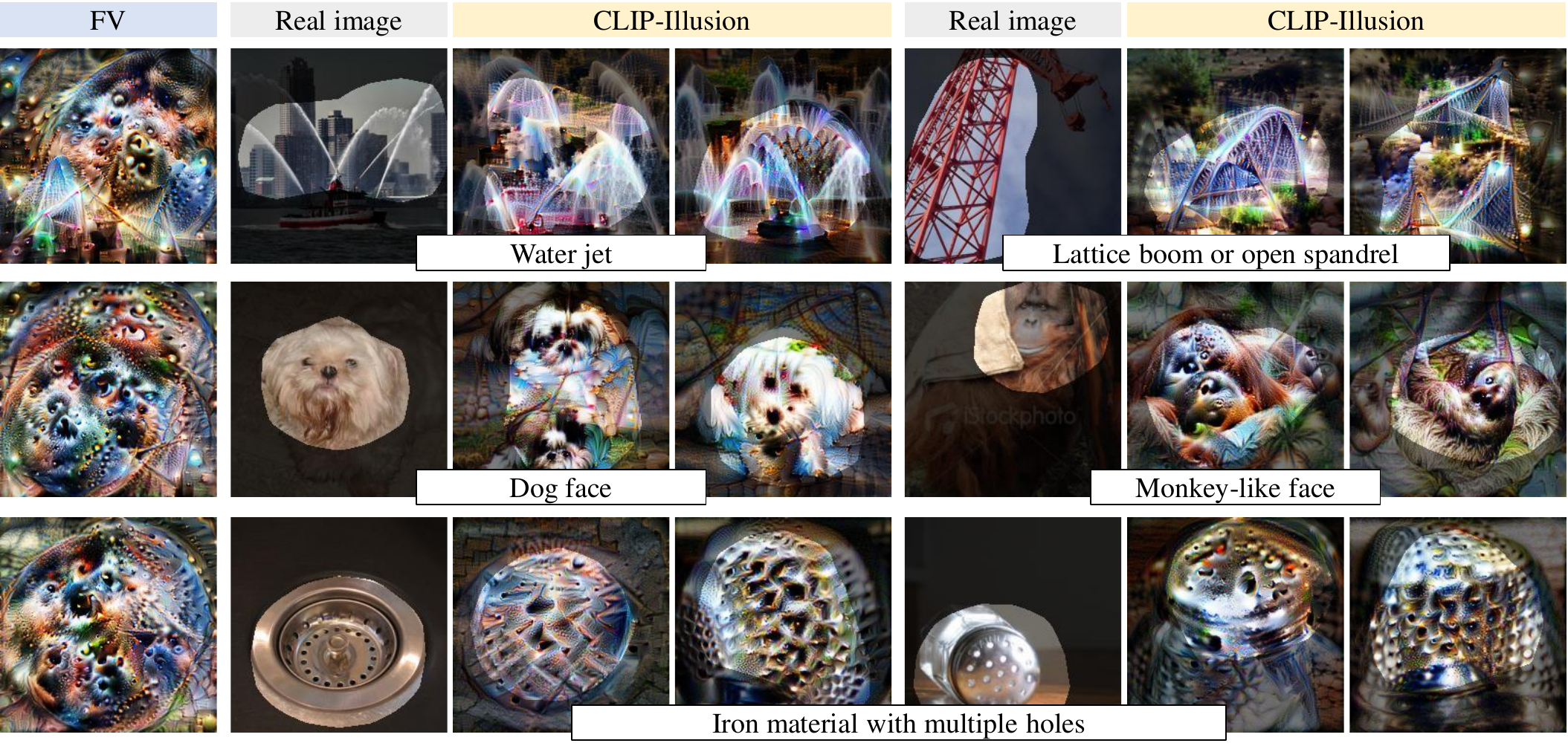}
\captionof{figure}{
    \textbf{Multiple facets of a given neuron.} The FV, real image, and CLIP-Illusion are visualizations for neuron 0 of the penultimate layer in ResNet-18. We chose CLIP-Illusion results corresponding to concepts of top-activated images, where CLIP-Illusion has automatically conditioned on 25 classes most related to neuron 0 based on $\beta$; The first row: fireboat, fountain, suspension bridge, and steel arch bridge. The second row: Shih-Tzu, Maltese, orangutan, and three-toed sloth. The third row: manhole cover, mike, saltshaker, and thimble. }
\label{fig:disentangle}
\end{figure*}

\subsection{Editing Decision Layer for Mitigating False Correlations}
Identifying the features admissible to mistakes in the model does not always lead to a straightforward solution, as DL practitioners may still find it challenging to resolve these issues even after affirming the underlying reasons for the mistakes. While the root causes (such as model scale and data noise) of errors in a DL model can vary, our focus is on identifying and mitigating false correlations that the model may learn.

Although we can directly lower the activation of discovered neurons or exclude them from predictions, these approaches can degrade the performance of the model because all classes refer to all features in $f$ in the dense layer. Therefore, we propose a method for editing the decision layer of the model to reduce the false correlations that may be learned from the training data. Because the importance of a particular feature in one class is determined by the extent to which the feature is considered relative to other classes, we minimize the influence of neurons that cause false correlations under a stochastic perspective to maintain the performance of the model as high as possible.

We determine the importance of each feature based on the extent to which it changed the probability of the class being investigated. In general classification networks, the logit for $i$-th class is calculated as follows:
\begin{equation}
    l_i=\sum^d_k {\beta_{i,k}\alpha_k} + b_i,
\end{equation}
where $\alpha_k$ is activation of $k$-th neuron and $\beta_{i,k}$ is $k$-th coefficient for class $i$. $b_i$ is a bias of $i$-th class. Then probabilities $P=[p_1,p_2,\cdots,p_c]$ for all classes are calculated by $\mbox{softmax}$ as follows:
\begin{equation}
\begin{aligned}
    p_i = {e^{l_i} \over {\sum^c_j {e^{l_j}}}} = {{\gamma_i e^{l'_i}} \over {\sum^c_j {\gamma_j e^{l'_j}}}}, \\
    \gamma_j=e^{\beta_{j,k}\alpha_k}, l'_j=l_j - \beta_{j,k}\alpha_k.
\end{aligned}
\end{equation}
To investigate how important the feature $\alpha_k$ is in the class $i$ compared to other classes, we calculate a changing ratio for probability when $\alpha_k$ is replaced to 0 as follows:
\begin{equation}
    {{p_i}\over{p'_i}} = \gamma_i \cdot{{{\sum^c_j {e^{l'_j}}}}\over {\sum^c_j {\gamma_j e^{l'_j}}}} = {{{\sum^c_j {e^{l'_j}}}}\over {\sum^c_j {e^{\alpha_k(\beta_{j,k}-\beta_{i,k})} \cdot e^{l'_j}}}},
\end{equation}
where $p'_i = e^{l'_i} / {\sum^c_j {e^{l'_j}}}$ and $l'$ are the probability and logit changed by the operation replacing $\alpha_k$ to zero, respectively. If we regularize the probability ratio to be equal to 1, it is equivalent to constraining $\alpha_k(\beta_{j,k}-\beta_{i,k})$ to be 0. The modified neurons are constrained to the extent of their weights in other classes. Thus, the higher the probability ratio, the more influential the feature $\alpha_k$ means. As $\beta_{j,k}-\beta_{i,k}$ is close to zero, the effect of the $k$-th feature on class $c$ decreases. From this perspective, we introduce a regularization term to minimize probability variation to mitigate false correlations by shrinking the effect of features activated through neurons on decision-making. To regularize the ratio of the two probabilities for neurons identified as the cause of false correlations, the decision layer is optimized by minimizing the following loss:
\begin{equation}
    \mathcal{L}_{edit} = \mathcal{L}_\text{CE}(y, h_t(h_b(x))) + \lambda_3 \left|{p_i / p'_i}-o\right|_2,
\end{equation}
where $o$ denotes a hyperparameter indicating the relative importance of the feature compared to the other classes. If $o$ is set to one, the decision layer is regulated such that class $c$ similarly considers the feature to other classes. $\lambda_3$ is a hyperparameter to balance $\mathcal{L}_\text{CE}$ and the regularization term. We retrain the decision layer of the model with the original training dataset by adding only the above regularization term. Our approach takes into account the correlations between the selected features and other classes. This stands in contrast to merely regularizing feature coefficients. The rationale behind this method is to manage the relative increase in correlations between features and other classes, thereby suppressing the occurrence of other false correlations. Our experiments show that our approach alleviates mistakes in the model by reducing false correlations. Notably, the proposed neuron-based debugging method does not require additional data collection.

\section{Experiments}

\subsection{CLIP-Illusion}
\textbf{CLIP-Illusion can disentangle concepts in a neuron.}  As illustrated in Fig.~\ref{fig:disentangle}, the top-activated images of a given neuron indicate that the neuron implies multiple concepts with even distant semantics. Thus, observing only a few images can convey data bias or misleading descriptions when interpreting the neurons. To show that CLIP-Illusion reveals multiple facets of the neuron without this drawback, we examine how a given neuron is expressed in each class by automatically conditioning the classes with the highest $\beta$ for the neuron in the decision layer. Class-conditional visualizations facilitate the interpretation of an abstract concept of neurons rather than leveraging probing datasets since our method finds the expression that maximizes that neuron within the class distribution learned by the model. We can label the concepts of neuron 0 of the penultimate layer in ResNet-18~\cite{he2016deep} pre-trained with the ImageNet dataset, as long diagonal lines, multiple round holes, and animal faces through visualizations, as shown in Fig.~\ref{fig:disentangle}.

\begin{figure*}[]
\centering
    \includegraphics[width=1.\linewidth]{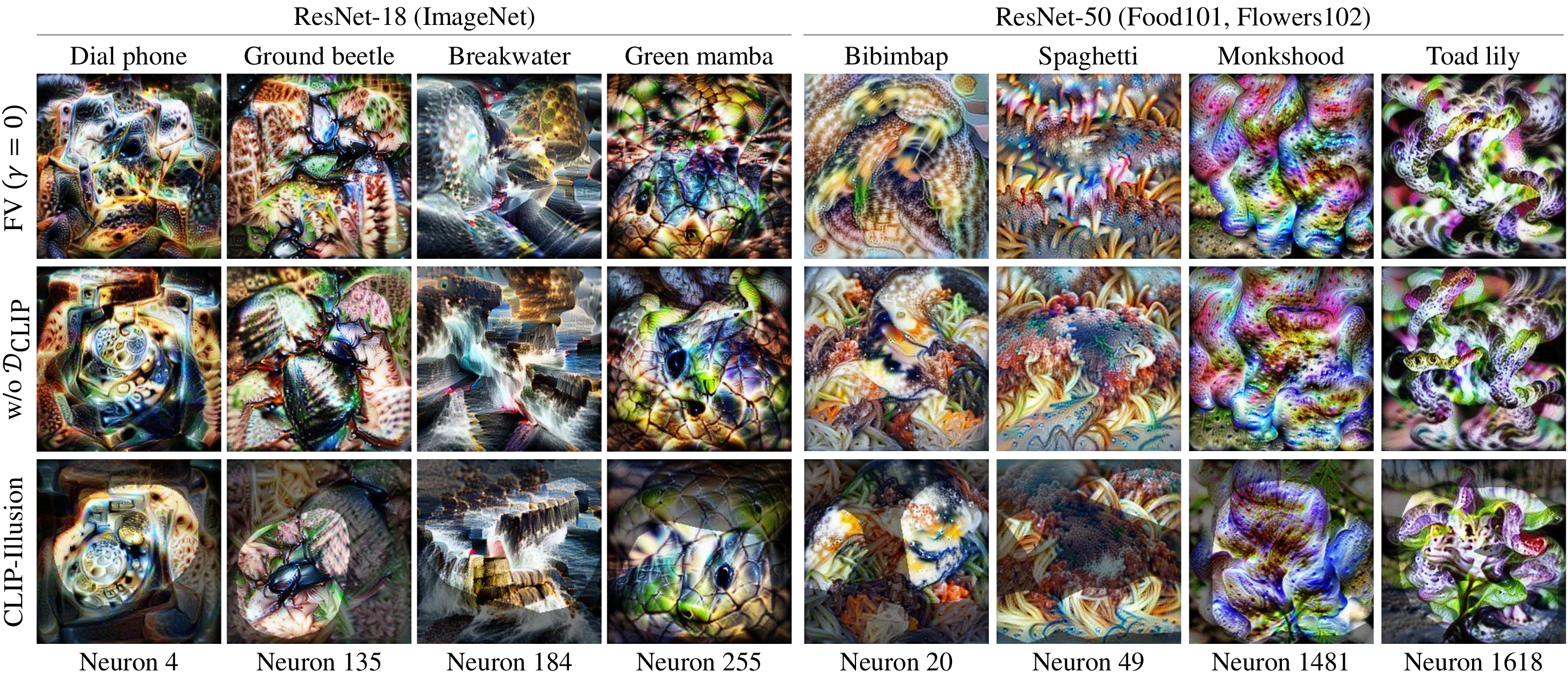}
\captionof{figure}{
    \textbf{Ablation studies of CLIP-Illusion.} The first and second rows are the results of ablation for our loss function. The captions above each column notate the conditional class name.}
\label{fig:abl}
\end{figure*}

\begin{table}[]
\caption{\textbf{Human evaluation for CLIP-Illusion.} The table presents the selected ratio of visualizations preferred by two groups: DL non-practitioners and DL practitioners specializing in computer vision. Visualizations were generated for 20 randomly chosen classes from the ImageNet 1000 classes using ResNet-18, employing three different methods: FV, CLIP-Illusion without CLIP loss, and CLIP-Illusion.}
\label{tab:human}
\begin{center}
\begin{small}
\begin{tabular}{lccc}
    \toprule
    \multirow{2}{*}{Groups} & \multicolumn{3}{c}{Selected ratio} \\
     & FV & w/o CLIP loss & CLIP-Illusion \\ \midrule
    Non-practitioners & 17.42\% & 39.89\% & \textbf{42.70\%} \\
    Practitioners & 3.75\% & 16.25\% & \textbf{80.0\%} \\
    \bottomrule
\end{tabular}%
\end{small}
\end{center}
\end{table}

\textbf{Ablation studies.} We investigate the effects of $\mathcal{D}_\text{CLIP}$ and class conditioning on the CLIP-Illusion through ablation studies. As shown in Fig.~\ref{fig:abl}, FV generates convoluted images that are difficult to interpret since the method represents multiple or abstract attributes of a neuron in a single image. Class-conditional visualizations are isolated into attribute directions associated with classes related to the neuron, thereby reducing complexity. Without the constraints of $\mathcal{D}_\text{CLIP}$, however, the images representing the concepts of the neurons in the classes remain challenging for humans to understand because they lack any constraints for activation. By contrast, CLIP-Illusion constrains visualizations with textual information, generating human-understandable visualizations that represent the attributes of the neuron in the class by creating class-centered images. Our experimental results show that CLIP-Illusion can represent neurons in a manner that is easy for humans to interpret, unlike visualizations that were abstracted or scattered throughout the image.

\textbf{Human Evaluation.}  To conduct experiments regarding human understanding of the CLIP-Illusion, we are first divided into two groups: a group of DL non-practitioners and a group of DL practitioners who study computer vision. We conducted surveys with 20 individuals from the general audience group and 8 individuals from the DL practitioners group. For our evaluation protocol, we selected a subset of 20 random classes from the ImageNet 1000 classes. Subsequently, we generated visualizations for the top neurons with the highest coefficients in the decision layer for each class in ResNet-18, employing CLIP-Illusion, CLIP-Illusion without CLIP loss, and FV. To ensure consistency, all visualizations for each method were generated as a single batch. To execute the evaluation for the general audience group, we leveraged Amazon Mechanical Turk (MTurk). Respondents were presented with a script requesting them to choose the most comprehensible visualization from the provided options as follows: ``The following images visualize one of the features related to class `\{class name\}.' Please choose a visualization that you find easy to understand from the images below.'' We identically applied the evaluation protocol for the group of DL practitioners. The selected visualization ratios are listed in Tab.~\ref{tab:human}. The human evaluation results clearly indicate the effectiveness of our CLIP-Illusion approach in terms of facilitating understanding. We provide more interpretable visualizations than the existing baseline for both groups, while our framework focuses on DL practitioners, as outlined in the Introduction Section.

\subsection{Mitigating Ground-Truth Spurious Correlations Learned from Synthesized Datasets}
We demonstrate the effectiveness of our neuron-based debugging approach in mitigating spurious correlations learned from synthesized datasets. We conduct experiments on two datasets: the MetaDataset~\cite{liang2022metashift} and the Waterbird dataset~\cite{2020Distributionally}. Our approach is explicitly incompatible with these explanations since we deal with neurons as the basis for mistakes in the model. Therefore, we conduct experiments to determine whether the identified neurons are responsible for the spurious correlation through changes in nominating concepts as the reasons for mistakes after modifying the discovered neurons.

\begin{figure}
\centering
    \includegraphics[width=1.0\linewidth]{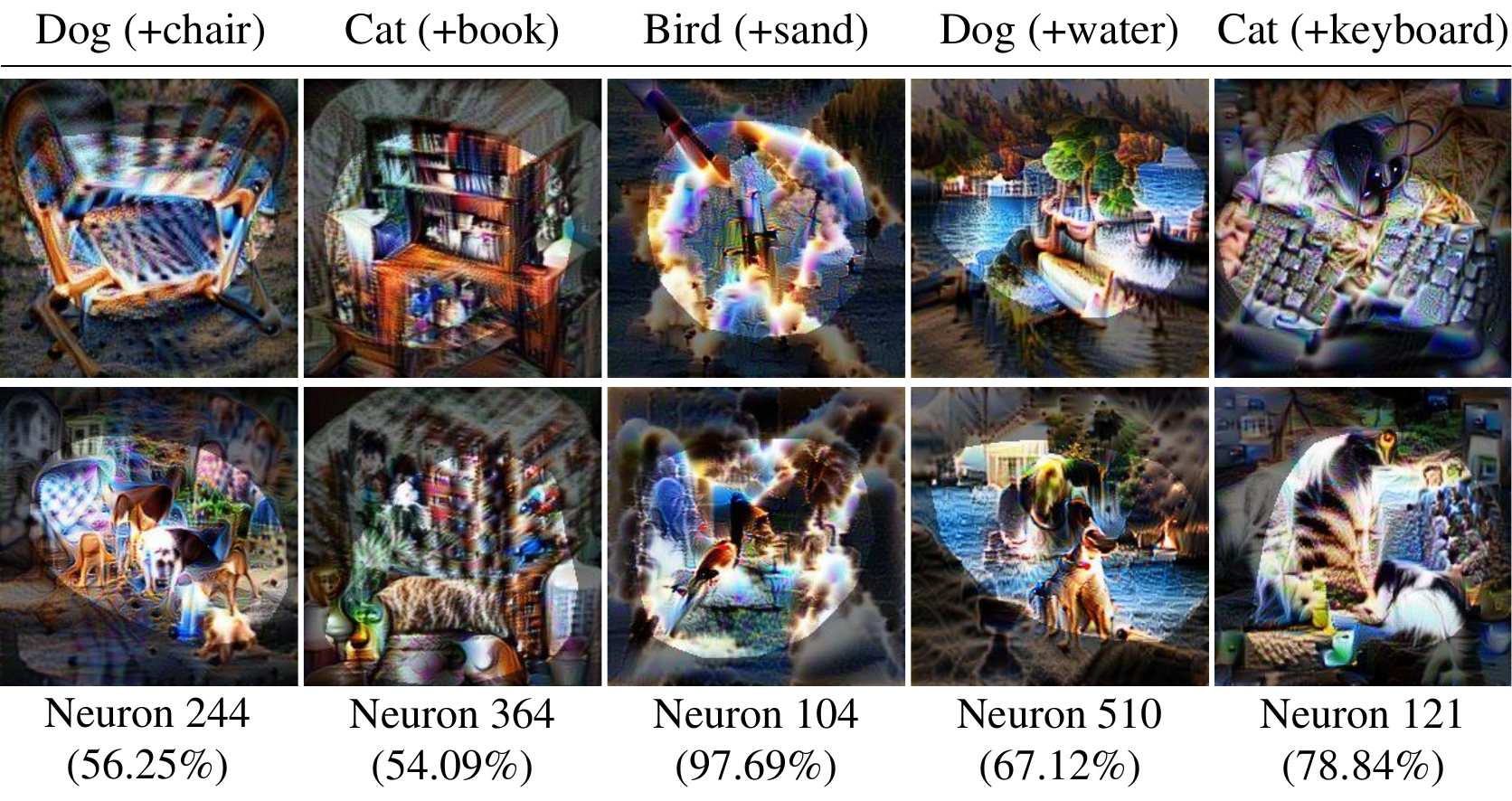}
    \caption{\textbf{Visualizations for features considered as spurious correlations in five scenarios.} This figure presents visualizations of CLIP-Illusion applied to neurons identified as responsible for spurious correlations. We selected these neurons based on features that exhibit lower correlations with the target class. The first row showcases visualizations generated using the original, untuned decision layer, while the second row features visualizations produced with the fine-tuned decision layer.}
    \label{fig:spur_v}
\end{figure}

\textbf{MetaDataset.} As in CCE~\cite{abid2022meaningfully}, we let the network learn the ground-truth spurious correlations to validate our method systematically. We construct five training scenarios where models classify five animal classes (cat, dog, bear, bird, and elephant). In each scenario, the models are trained on a dataset where all training images within a class contain a specific attribute and then tested on images of the class without the attribute. The trained models in all training scenarios achieved an accuracy of at least 70\% for the validation set in the scenarios. To explicitly investigate the effect of our editing approach on our framework, we leverage CCE, which utilizes a concept bank to identify spurious correlations. The CCE measures concept scores for the causes of mistakes in the model based on the concept bank, where the concept score is the weight of the vector of the concept to be added to correct incorrect predictions. To employ the constructed concept bank of CCE, we only tuned the decision layer of a ResNet-18 network pre-trained on the ImageNet dataset. We evaluate the effectiveness of the editing approach by measuring the changes in accuracy, precision@3, and concept scores before and after editing. Furthermore, we compare the model edited under our stochastic view with the edited model (CON) by constraining the coefficients for the activation of neurons by minimizing the following loss:
\begin{equation}
    \mathcal{L}_{CON} = \mathcal{L}_\text{CE}(y,h_t(h_b(x))) + \left|\alpha_k\cdot\beta_{i,k} + b_i\right|_2,
\end{equation}
where $b_i$ denotes a bias of $i$th class. $o$ in $\mathcal{L}_{edit}$ is set to 1 for all scenarios. As illustrated in Fig.~\ref{fig:spur_v}, our neuron-based debugging method determines the spurious correlations with high-level concepts across multiple neurons after observing a small number of neurons. Tab.~\ref{tab:edit} demonstrates that the proposed approach successfully reduces model errors, mitigating spurious correlation. Conversely, although the CON approach meaningfully alleviates the ground-truth spurious correlations, it degrades the performance of the model. These results indicate that the influence of features should be determined by considering the weights of the features in other classes.

\begin{figure}
\centering
    \includegraphics[width=1.0\linewidth]{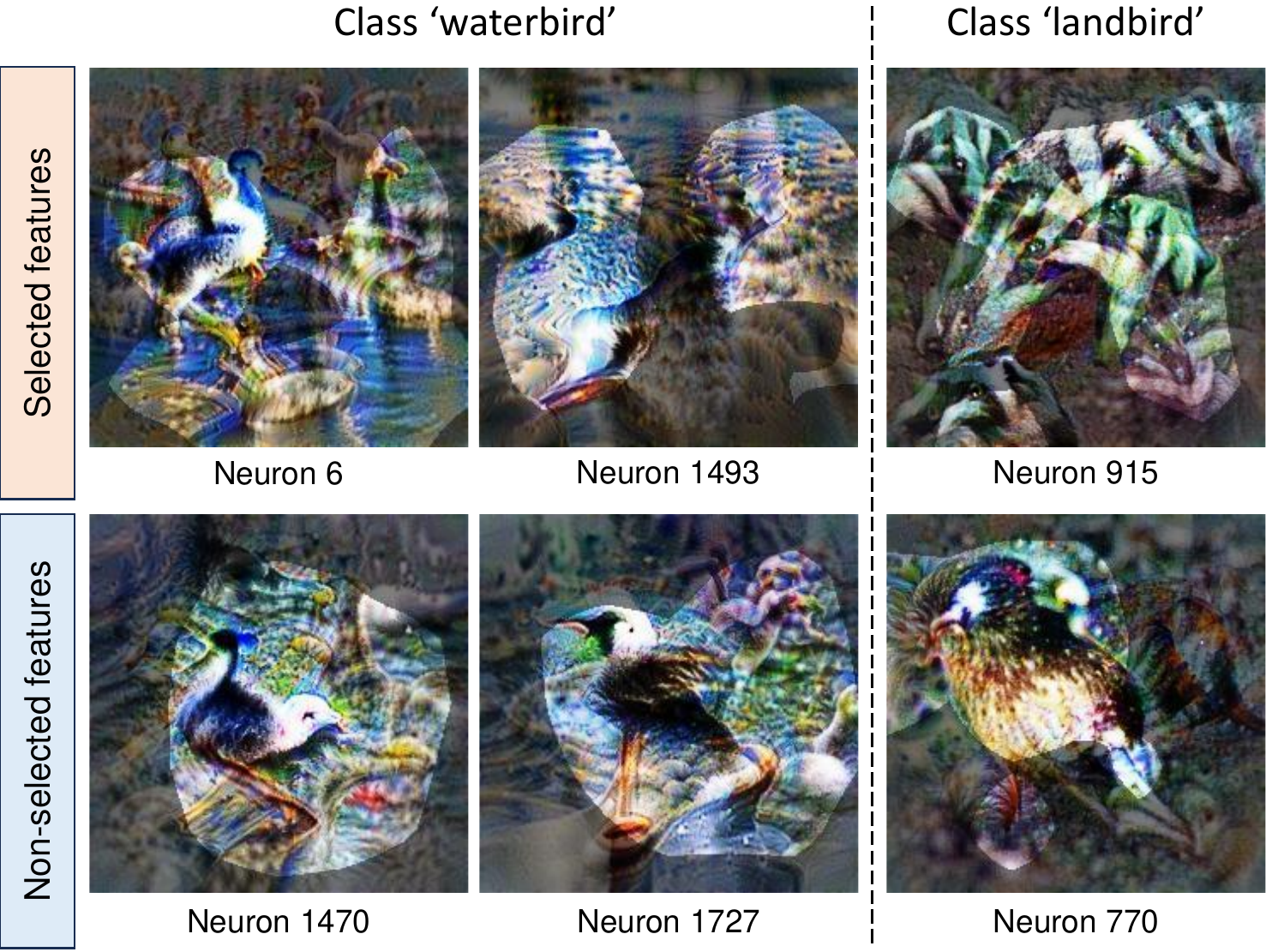}
    \caption{\textbf{Visualizations for features considered as spurious correlations learned from the Waterbird dataset.} This figure shows the visualizations of CLIP-Illusion for neurons identified as causing spurious correlations, such as water and forest.}
    \label{fig:waterbird}
\end{figure}

\textbf{Waterbird.} The Waterbird dataset consists of images of birds categorized as either `waterbirds' or `landbirds.' The images are taken against different backgrounds, either water or land, which creates a visual context for the birds. The four group labels in this dataset reflect the environmental context (water or land) and the alignment between the bird category and the background. In the training set of the Waterbird dataset, 95\% of the waterbirds are placed against a water background, whereas the remaining 5\% are placed against a land background. Similarly, 95\% of the landbirds are placed against a land background, and the remaining 5\% are placed against water. We fine-tune a pre-trained ResNet-50 network for the dataset via empirical risk minimization (ERM). Fig.~\ref{fig:waterbird} shows the spurious correlations (water and forest) indicated by 6 neurons chosen after probing visualizations for 30 of the 2048 neurons. In this experiment, we simultaneously constrain the identified neurons for both classes. We also compare our approach to two other methods: DISC~\cite{wu23disc} and JTT~\cite{liu2021just} to mitigate spurious correlation without group labels. In Tab.~\ref{tab:waterbird}, the DISC method effectively alleviates spurious correlations, particularly for the worst-performing group. However, they employ external data, specifically a concept dataset, and utilize mixup augmentation. The JTT method also improves the performance of the worst-performing group, but it incurs a significant drop in overall accuracy. In contrast, our proposed method exhibits a robust trade-off by simultaneously improving the worst-group performance and maintaining a high overall accuracy. Notably, when enhanced with checkpointing by class accuracy, our method performs competitively with existing methods in terms of worst-group accuracy, while still maintaining an above-average accuracy over all groups. Moreover, our strategy of visualizing the activations of specific neurons provides an interpretable debugging approach, enhancing the transparency and understanding of the bias mitigation process.

\subsection{Improving the Performance of Worst Class in Fine-grained Classification}
In real-world settings, we cannot clearly determine the false correlations learned by the model. In addition, it is challenging to explicitly represent concepts learned by the model in unique domains such as fine-grained image classification. In this section, we conducted experiments to demonstrate the effectiveness of our framework in improving the performance of the worst classes in fine-grained classification, even in real-world settings, where the ground-truth false correlation is not always clear. To alleviate class imbalance, we identify false correlations in the worst class with the lowest accuracy for ResNet-50 networks learned from two datasets: Flowers102~\cite{nilsback2008automated} and Food101~\cite{bossard2014food}.

\begin{table*}[]
\caption{\textbf{Mitigating Spurious Correlations by Editing the Decision Layer.} The $\vartriangle$ denotes the amount of change from measures for the original model, with positive values indicating an increase and negative values indicating a decrease. The term Acc refers to the accuracy of the model, thus a higher $\vartriangle$ Acc indicates an improved regularization effect and a decrease in the error of the model. Prec@3 refers to the precision rate of the top-3 ranked reasons for mistakes for the target spurious concept. Therefore, lower values of $\vartriangle$ Prec@3 and $\vartriangle$ Concept score indicate a greater alleviation of the target spurious correlation. }
\label{tab:edit}
\begin{center}
\begin{small}
\begin{tabular}{lccccccccc}
\toprule
\multicolumn{1}{c}{Class (+correlation)} & \multicolumn{1}{c}{\% Acc} & \multicolumn{1}{c}{\% Prec@3} & \multicolumn{1}{c}{Concept score} & \multicolumn{2}{c}{\% $\vartriangle$ Acc $\uparrow$} & \multicolumn{2}{c}{\% $\vartriangle$ Prec@3 $\downarrow$} & \multicolumn{2}{c}{$\vartriangle$ Concept score $\downarrow$} \\
\multicolumn{1}{c}{Method} & \multicolumn{3}{c}{-} & \multicolumn{1}{c}{CON} & \multicolumn{1}{c}{Ours} & \multicolumn{1}{c}{CON} & \multicolumn{1}{c}{Ours} & \multicolumn{1}{c}{CON} & \multicolumn{1}{c}{Ours} \\ \midrule
Dog (+chair)    & 49.35 & 81.7 & 0.492  & -2.74  & \textbf{+12.37} & -74.4 & \textbf{-78.0}  & -0.052 & \textbf{-0.162} \\
Cat (+book)     & 67.61 & 93.5 & 0.526  & -4.09  & \textbf{+2.65}  & -41.4 & \textbf{-53.0}  & -0.1   & \textbf{-0.113} \\
Bird (+sand)    & 58.51 & 94.4 & 0.589  & -6.53  & \textbf{+19.27}  & -4.4  & \textbf{-5.1}   & -0.1 & \textbf{-0.108} \\
Dog (+water)    & 39.22 & 85.0 & 0.642  & \textbf{+4.65}  & +3.22  & -6.9  & \textbf{-43.3}  & -0.077 & \textbf{-0.108} \\
Cat (+keyboard) & 54.53 & 38.3 & 0.432  & +7.66  & \textbf{+23.29}  & -11.4  & \textbf{-30.3} & -0.076 & \textbf{-0.098} \\ \bottomrule
\end{tabular}%
\end{small}
\end{center}
\vskip -0.1in
\end{table*}

\begin{table}[]
\caption{\textbf{Experimental comparison of methods for mitigating spurious bias learned from the Waterbird dataset.} The results underline the effectiveness of different approaches in mitigating spurious bias. (C) refers to the result of checkpointing based on the minimum class accuracy. }
\label{tab:waterbird}
\begin{center}
\begin{tabular}{lccc}
\toprule
\multicolumn{1}{l}{\multirow{2}{*}{Method}} & \multicolumn{3}{c}{Waterbirds} \\
\multicolumn{1}{c}{} & Avg. Acc. & Worst class Acc. & Worst group Acc.\\ \midrule
ERM  & 91.73\% & 79.83\% & 66.04\%\\
\midrule
ERM+DISC~\cite{wu23disc} & 89.78\% & 89.38\% & 83.64\%\\ \midrule
ERM+JTT~\cite{liu2021just}  & 86.76\% & 83.88\% & 71.50\% \\
ERM+Ours & 91.35\% & 81.31\% & 68.38\% \\
ERM+Ours (C) & 90.65\% & 83.88\% & 72.43\% \\ \bottomrule
\end{tabular}
\end{center}
\end{table}

\begin{figure}
\centering
    \includegraphics[width=1.0\linewidth]{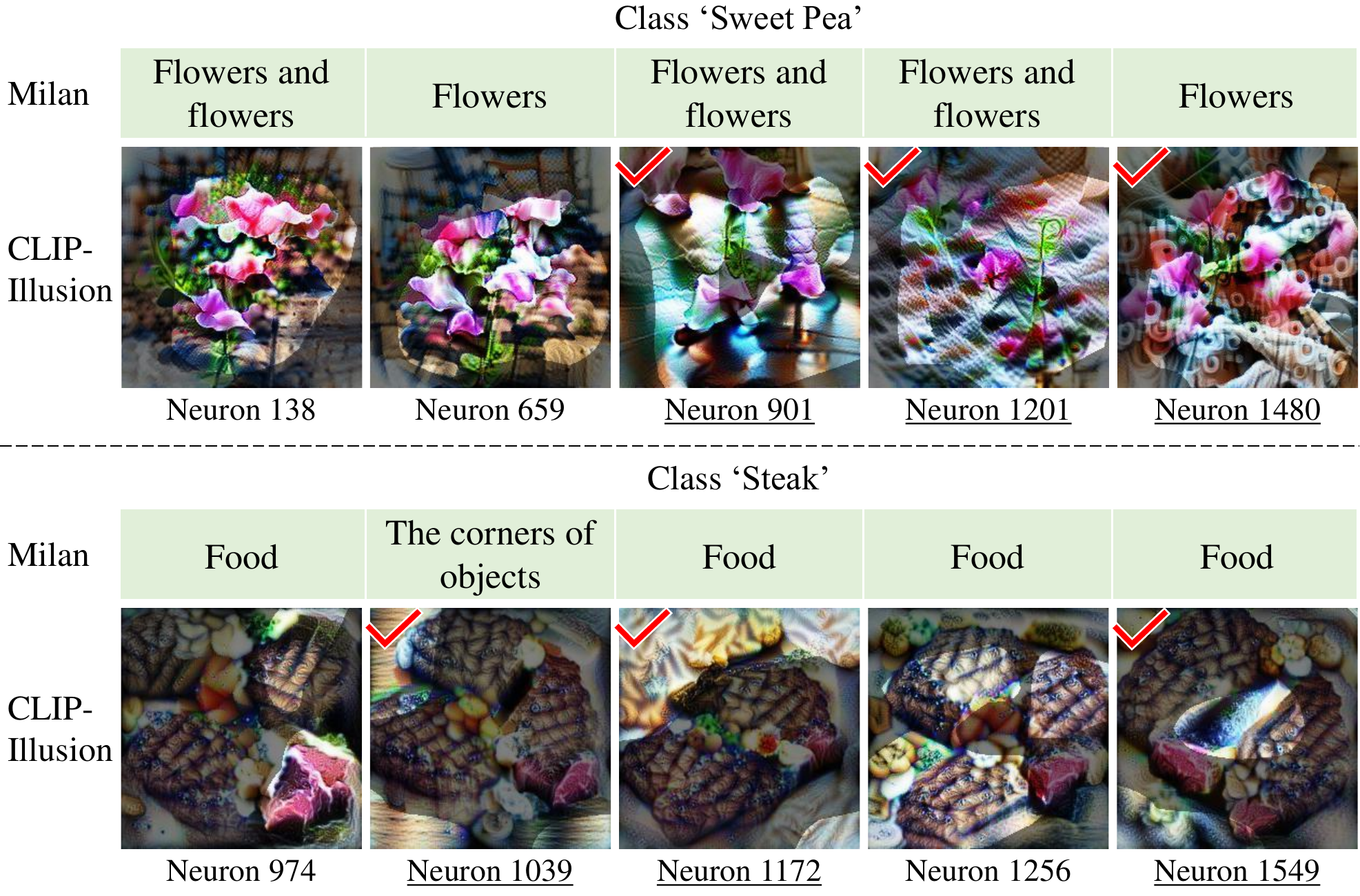}
    \caption{\textbf{Visualizations for features conditioned on worst classes in fine-grained classification.} The images with red checks are features designated as false correlations in the target class. The first row and second row show visualizations for neurons within networks trained on Flowers102 and Food101, respectively.}
    \label{fig:hard}
\end{figure}

\begin{table*}[]
\caption{\textbf{Experimental comparison of methods for improving worst class in fine-grained classification.} The table shows the results of an experiment to improve the performance of hard classes in fine-grained classification on two datasets: Flowers102 and Food101. We compare our debugging method with class weighting and focal loss, which weigh certain classes. The ``Class Acc'' and ``Avg Acc'' mean accuracy for the target hard class and mean accuracy for total samples, respectively. The ``Min Class Acc'' represents the lowest accuracy among all the classes.}
\label{tab:hard_class}
\vskip 0.05in
\begin{center}
\begin{small}
\begin{tabular}{lcccccc}
\toprule
Datasets         & \multicolumn{3}{c}{Flowers102}   & \multicolumn{3}{c}{Food101} \\ \midrule
Methods          & Class Acc        & Avg Acc       & Min Class Acc    & Class Acc     & Avg Acc   & Min Class Acc\\ \midrule
Original         & 33.33\%          & 84.66\%       & 33.33\% (3)      & 43.2\%           & 82.65\%     &  43.2\% (93) \\ 
Class weighting & 33.33\%          & 84.72\%       & 33.33\% (3)      & 53.6\%           & \textbf{82.99\%}     &  53.6\% (93) \\
Focal loss~\cite{lin2017focal}     & 33.33\%   & \textbf{84.76\%}    & 33.33\% (3)   & 39.6\%     & 82.69\%     &  39.6\% (93) \\
Ours             & \textbf{58.33\%} & 84.26\%  & \textbf{49.58\%} (50)   & \textbf{58.0\%}  &  82.20\%    &  \textbf{54.0\%} (8) \\ \bottomrule
\end{tabular}%
\end{small}
\end{center}
\vskip -0.1in
\end{table*}

\textbf{Datasets.} The Food-101 dataset comprises a collection of food images categorized into 101 classes. Each class contains 1,000 images, with 250 images allocated to the test set and 750 images for the training set. Because the dataset does not provide a separate validation set, we extracted a validation set by randomly selecting 15\% of the training data for each class. The Flowers-102 dataset contains images of flowers categorized into 102 classes. The dataset exhibits a class imbalance issue since each class is composed of a varying number of images, ranging from 40 to 258, with an average of 77.89 images per class.

For each model, we manually chose three neurons after observing visualizations of neurons ranked in the top 30 as underlying reasons. As illustrated in Fig.~\ref{fig:hard}, Milan, a text decoding method for neuron interpretation~\cite{hernandez2021natural}, provides abstract labels when inferring with a base model without domain-specific training data. In contrast, CLIP-Illusion can identify neurons that embed features not directly related to the class through visualization, emphasizing the utilization of visualizations. For instance, we reveal that neuron 1480 encompasses features like the watermark present in data for the sweet pea class within the Flowers102 dataset. Following this, we improve inferences for mistakes by editing the decision layer without any additional training data or prior knowledge. We empirically set $o$ to $\frac{1}{3}(p/p'-1) + 1$ calculated with the lowest probability ratio of neurons to be edited after calculating $p_i/p'_i$ for each neuron $i$ from the mean features of the input samples. Our approach led to a significant improvement in the accuracy of the target class compared to other methods of addressing class imbalance while maintaining reasonable average accuracy for all test samples, as presented in Tab.~\ref{tab:hard_class}. The results obtained from the Flowers-102 dataset demonstrate the superiority of our framework. Despite the inherent class imbalance in the dataset, our framework effectively improves the performance of the worst class without significantly compromising the overall performance. Our method presents controllable debugging at the feature level while the performance of the model is somewhat degraded. This experiment highlights the contributions of our interpretable framework in improving performance in real-world scenarios.

\section{Limitations}
\begin{figure}[]
\centering
\includegraphics[width=1.\linewidth]{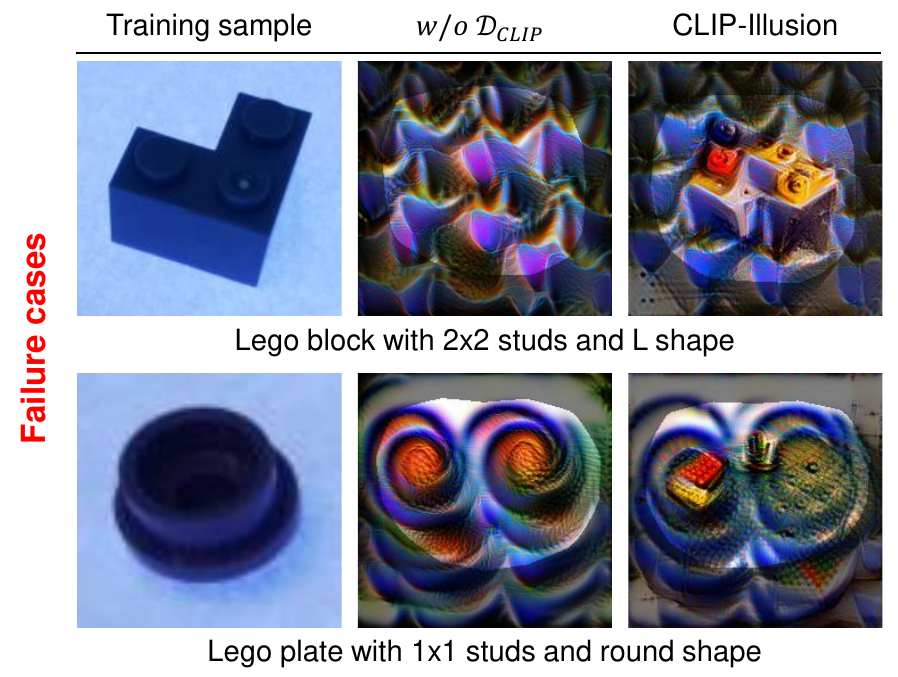}
\captionof{figure}{
    \textbf{Failure cases of CLIP-Illusion due to domain misalignment.} The figure represents visualizations for neurons of the penultimate layer in ResNet-18 trained on the Lego brick dataset by conditioning the most relevant classes written as below captions. }
\label{fig:hyper_d}
\end{figure}

First, the utilization of CLIP networks introduces two limitations: domain misalignment and dependence on prompt design. For instance, visualizations generated from a prompt for the class "Image of a cornet, horn, trumpet, trump" may inadvertently include the face of former US President Trump alongside a musical instrument. We alleviate this issue by disregarding regions that are less relevant to the properties of neurons. In the loss function $\mathcal{L}_\text{CI}$, $\epsilon$ is a hyperparameter introduced to prevent domain misalignment between the CLIP network and the probed network. Despite the activation of these neurons, there may be a failure to generate visualizations that humans can interpret due to a mismatch between the CLIP network and the learned class information. As shown in Fig.~\ref{fig:hyper_d}, we conducted an experiment applying CLIP-Illusion to a Lego brick classifier trained with a Lego brick dataset~\cite{lego_dataset} that is out-of-domain of the CLIP network. In this out-of-domain scenario, the results of CLIP-Illusion may not be directly interpretable by humans, leading to a limitation of our framework. However, the constraints applied by the CLIP loss still guide in comprehending the underlying representations learned by the Lego block classifier.

Second, our debugging framework relies on the ability of DL practitioners to identify neurons while recognizing causes for mistakes in the model regardless of any domain. The framework does not provide direct guidance but operates similarly to general code debuggers by indicating the location of the causes of the errors, thereby requiring human intervention. Thus, we considered providing slight guidance to neurons corresponding to spurious features through core relevance scores to aid the interpretation of the practitioners. The core relevance score quantifies the relevance of features activating a neuron $n$ in the context of core features of a class $c$. We calculated the score by comparing the cosine similarity between the feature vector for visualization of neuron $n$ and the class representative vector before and after mask removal. We confirmed that neuron selection through the core relevance score matched human selection in cases where the class and the feature generally had a low correlation (\textit{i.e.}, our experiments for fine-grained classification in the real world). However, we discovered that core relevance scores cannot be applied when a class, such as `waterbird,' and a feature, such as `water,' naturally exhibit high correlations but lack causality since they also show high correlations in the validation set.

\section{Conclusion}
In this study, we proposed NeuroInspect, a neuron-based debugging framework for deep learning models. NeuroInspect aims to uncover the underlying reasons for the mistakes made by models and provide human-understandable explanations without requiring additional data or modifying the decision-making process. Our framework utilizes counterfactual explanations to identify neurons responsible for mistakes and a feature visualization approach to translate them into visualizations that are easy for humans to interpret. We introduced CLIP-Illusion, a class-conditional feature visualization method that enhances the interpretability of visualizations by considering the connections between the features and the decision layer. CLIP-Illusion overcomes the limitations of existing visualization methods by generating images that instantiate abstract and mixed features into specific features expressed in a specific class. We conducted experiments to evaluate the effectiveness and practicality of our framework. We demonstrated that NeuroInspect can identify and alleviate spurious correlations learned from synthesized datasets, leading to improved model performance and more reliable decision-making. We also showed that our framework can improve the performance of the worst-performing classes in fine-grained classification tasks, even in real-world settings where ground-truth false correlations are not always clear. Overall, our approach provides a practical model debugging method that can help DL practitioners understand and interpret the decision-making processes of their models. Our framework can be deployed without requiring significant changes to the model or additional data, making it a valuable tool for improving the transparency and accountability of deep learning models.

\vspace{11pt}

\vfill

\bibliographystyle{IEEEtranN}
\bibliography{main}

\newpage
\appendix
\section{Appendix}

\subsection{CLIP-Illusion}

\noindent\textbf{Implementation details.} In this section, we provide details for hyperparameters and augmentations for FV and CLIP-Illusion. We initialize the image $I_{opt}$ being optimized from Gaussian noise and then optimize the image to maximize activation for a given neuron. As in \citet{olah2017feature}, we apply rotation, translation, and scaling transformations when optimizing the image. By the transformations, the image is rotated by a random angle within -15 degrees to 15 degrees, randomly moved in vertical and horizontal directions within 15\% of the image size, and scaled by a random factor within the range of 0.7 to 1.2. Additionally, Gaussian smoothing~\cite{cohen2019certified} is applied for better visualization. We create visualizations by optimizing the image $I_{opt}$ using AdamW~\cite{loshchilov2018decoupled} optimizer and Cosine Annealing~\cite{loshchilov2017sgdr} learning rate scheduler, where the initial learning rate is set to $9\times10^{-3}$. Our all experiments are conducted with a single A6000 GPU. We describe settings for hyperparameters of CLIP-Illusion in the following sections. Please refer to our code for detailed experimental configurations.

\noindent\textbf{Prompt design.} We use prompts applied with several formats to condition class information. We follow formats of CLIP~\cite{radford2021learning} for zero-shot image classification. Thus, a conditioning text embedding corresponds to the mean of embeddings for texts in multiple formats.

\begin{figure*}[]
\centering
\includegraphics[width=1.\linewidth]{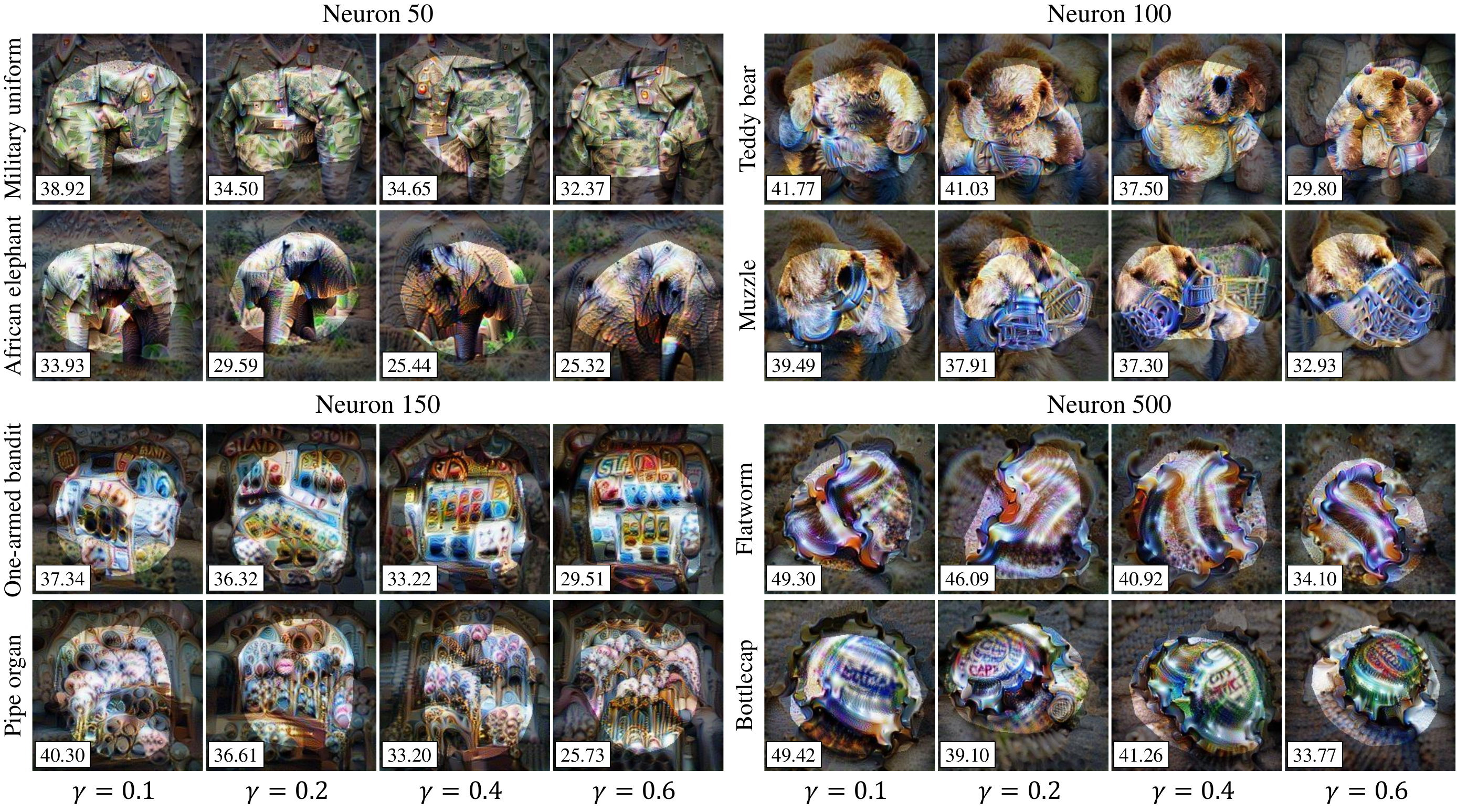}
\captionof{figure}{
    \textbf{Results of CLIP-Illusion over $\gamma$.} The visualizations were generated for neurons of the penultimate layer in ResNet-18 by conditioning the most relevant classes written as left captions. The values at the bottom left of the image represent activation values for neurons. We set $\epsilon$ and the number of optimizations to 0.1 and 400, respectively. }
\label{fig:hyper_r}
\end{figure*}

\begin{figure*}[]
\centering
\includegraphics[width=1.\linewidth]{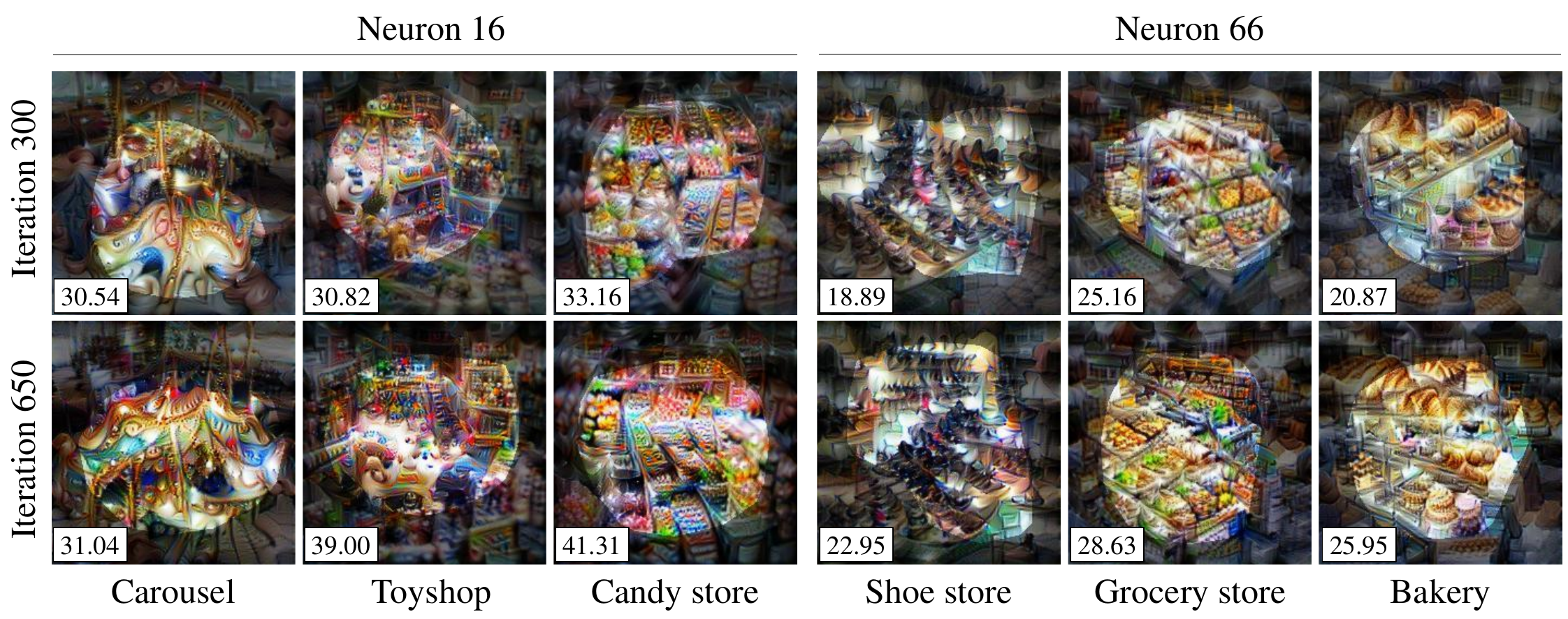}
\captionof{figure}{
    \textbf{Results of CLIP-Illusion over the number of optimizations.} The visualizations were generated for neurons of the penultimate layer in ResNet-18 by conditioning the most relevant classes written as below captions. The values at the bottom left of the image represent activation values for neurons. }
\label{fig:hyper_it}
\end{figure*}

\noindent\textbf{Experiments for hyperparameters.} We investigate the tendency of CLIP-Illusion over hyperparameters in this section. The final loss of CLIP-Illusion is as follows:
\begin{equation}
    \mathcal{L} = -(\mathcal{D}_\text{CLIP}+\epsilon)\cdot(\alpha_n + \gamma l_c),
\end{equation}
where $\gamma$ was introduced to control class representations. A higher value of $\gamma$ leads to stronger class-specific features, as multiple features are activated to represent the classes. Consequently, the activations of the probed neurons may decrease since the class representations become more prominent. In Fig.~\ref{fig:hyper_r}, when $\gamma$ is small, even though the activations were conditioned on different classes, they exhibit similar features. As $\gamma$ increases, the class-specific features become more pronounced, and the visualizations align more closely with the characteristics of the respective classes. While the activation of the neurons decreases to some extent with higher $\gamma$, the class-centric visualizations clearly illustrate the specific features through which the neurons are expressed in each class, facilitating human interpretation.

To generate visualizations by minimizing the loss, we repeatedly optimize images initialized from Gaussian noise over a fixed number of iterations. We investigated the impact of the number of iterations on the activation and fidelity of the generated visualizations. By varying the number of optimization steps, we aimed to understand how the additional iterations affect activations and the level of clarity in the visual representations. The visualizations obtained after 300 steps of optimization provide initial insights into the activated features and patterns associated with the probed neurons. These visualizations may still exhibit some level of noise and lack fine-grained details. On the other hand, extending the optimization process to 650 steps allows for further refinement of the visualizations. The additional iterations enable the model to capture more intricate features and enhance the overall activations of the generated representations. As a result, the visualizations become more distinct and easier to interpret, providing a clearer understanding of the specific features expressed by the probed neurons for each class. Fig.~\ref{fig:hyper_it} presents a visual comparison between the visualizations obtained at 300 and 650 steps of optimization, showcasing the progressive improvement in the quality and level of detail achieved with the increased number of iterations. However, it is important to consider the average activations of the probed model and the time required for optimization when determining the appropriate number of iterations for generating visualizations. Tab.~\ref{tab:it} presents the average activations and average required time of the visualizations based on the number of iterations.

We also discuss the effects of the $\epsilon$ with the out-of-domain problem in the following section.

\begin{table*}[]
\caption{\textbf{Average activations and required time of CLIP-Illusion.} The table compares the average activations and average required time for results of CLIP-Illusion generated for all neurons in the penultimate layer of ResNet-18 pre-trained on ImageNet, according to the number of iterations. All visualizations were generated using a batch size of 3. The values are compared with the activations of the top-3 activated real images on ImageNet validatation set.}
\label{tab:it}
\vskip 0.05in
\begin{center}
\begin{small}
\begin{tabular}{cccc}
\toprule
Visualization & Iteration & Average activation & Average time \\ \midrule
\begin{tabular}[c]{@{}c@{}}Top-3 activated images\end{tabular} & - & 25.21 & - \\ \midrule
\multirow{4}{*}{CLIP-Illusion} & 300 & 30.58 & 12.33 sec \\
 & 400 & 32.91 & 16.45 sec \\
 & 500 & 34.47 & 21.82 sec \\
 & 650 & 36.13 & 26.49 sec \\ \bottomrule
\end{tabular}
\end{small}
\end{center}
\vskip -0.1in
\end{table*}

\subsection{Neuron-based Debugging Framework}

\noindent\textbf{Finding neurons causing false correlations.} Spurious correlations are explicitly not revealed in the real world. Therefore, rather than targeting concepts that are spurious correlations, we find neurons with concepts that are less related to the class to be investigated. Across all target samples, we acquire top-5 ranked neurons from $\omega$ optimized via counterfactual explanation for a single sample. Then, we interpret concepts of the neurons with a top-5 rank ratio of 3\% or more through CLIP-Illusion.

\noindent\textbf{Experiments for ground truth spurious correlations.} The MetaDataset utilized in the experiment for counterfactual explanation consists of labeled datasets of animals in different settings and with different objects. We use 100 training samples and 100 validation samples in each training scenario. The number of test samples in each scenario "Dog (+chair)", "Dog (+water)", "Cat (+book)", "Bird (+sand)", and "cat (+keyboard)" is 2401, 2302, 1661, 1562, and 1632, respectively. We conduct fine-tuning ResNet-18 by optimizing a decision layer with the Adam optimizer and the Cosine Annealing learning rate scheduler. A learning rate, batch size, and the number of epochs are set to 0.002, 8, and 20, respectively. In the stage of editing, $\lambda_3$, the learning rate, and batch size are set to 1.0, 0.001, and 16. We employ the best model based on the accuracy of the validation dataset. We adopt SGD as an optimizer for optimization in counterfactual explanation. In the editing stage for the Waterbird dataset, we set a learning rate, the number of epochs, $\lambda_3$, and $o$ to $5\cdot10^{-5}$, 20, 0.001, and 1.0, respectively. Similar to baselines, we use the Adam optimizer with a weight decay of 0.0001.

\noindent\textbf{Experiments for fine-grained image classifiers in real-world settings.} The Food-101 dataset consists of a collection of food images categorized into 101 classes. Each class contains 1,000 images, with 250 images allocated for the test set and 750 images for the training set. Since the dataset does not provide a separate validation set, we extracted a validation set by randomly selecting 15\% of the training data for each class. For our experiments, we utilized the ResNet-50 model with the Adam optimizer and the Cosine Annealing learning rate scheduler~\cite{loshchilov2017sgdr}. The training process was performed using a batch size of 128 and an initial learning rate of 0.002. We trained the model for 20 epochs and saved the checkpoint with the best accuracy on the validation set.

The Flowers-102 dataset contains images of flowers categorized into 102 classes. Each class is composed of a varying number of images, ranging from 40 to 258 images. The dataset comprises a total of 6,387 images, with an average of 77.89 images per class. The training settings for the Flowers-102 Dataset is the same as those used for the Food-101 dataset.

For neuron-based editing for decisions, we employed the same batch size, optimizer, and number of epochs as in the original training stage. We adopted a warmup scheduler with a maximum learning rate of 0.001. During training, we applied early stopping if the best accuracy for a validation set did not improve for five consecutive updates. The $o$ in $\mathcal{L}_{edit}$ for the Food-101 classifier and the Flowers-102 classifier were set to 1.02 and 1.003, respectively. The hyperparameter $\lambda_3$ is set to 0.01.

\end{document}